\documentclass[conference]{IEEEtran}
\IEEEoverridecommandlockouts
\usepackage{cite}
\usepackage{amsmath,amssymb,amsfonts}
\usepackage{algorithmic}
\usepackage{graphicx}
\usepackage{textcomp}
\usepackage{xcolor}
\def\BibTeX{{\rm B\kern-.05em{\sc i\kern-.025em b}\kern-.08em
    T\kern-.1667em\lower.7ex\hbox{E}\kern-.125emX}}

\usepackage{amsthm}
\theoremstyle{plain}
\newtheorem{definition}{Definition}
\newtheorem{corollary}{Corollary}

\usepackage{xcolor}
\usepackage{mdframed}

\usepackage{bm}
\usepackage{url}

\usepackage[flushleft]{threeparttable}
\usepackage{booktabs}
\usepackage{tabularx}
\usepackage{multirow}
\usepackage{makecell}
\usepackage{enumitem}
\usepackage{hyperref}
\usepackage{subcaption}
\usepackage{pgf,pgfplots}
\usetikzlibrary{fit,calc}
\usepgfplotslibrary{external}
\usepackage{tikz}

\begin{document}
\title{
A User Study on Contrastive Explanations for Multi-Effector Temporal Planning with Non-Stationary Costs
\thanks{This work has been submitted to the IEEE for possible publication. Copyright may be transferred without notice, after which this version may no longer be accessible. This work was partially funded by EPSRC CHAI project (EP/T026707/1) and EPSRC IAA fund at the University of Bristol. Code is available online: \url{https://github.com/kevinmcareavey/cuttlefish-setup}.}
}

\author{
\IEEEauthorblockN{Xiaowei Liu}
\IEEEauthorblockA{\textit{University of Bristol} \\
Bristol, UK \\
\href{mailto:xiaowei.liu@bristol.ac.uk}{xiaowei.liu@bristol.ac.uk}
}
\and
\IEEEauthorblockN{Kevin McAreavey}
\IEEEauthorblockA{\textit{University of Bristol} \\
Bristol, UK \\
\href{mailto:kevin.mcareavey@bristol.ac.uk}{kevin.mcareavey@bristol.ac.uk}
}
\and
\IEEEauthorblockN{Weiru Liu}
\IEEEauthorblockA{\textit{University of Bristol} \\
Bristol, UK \\
\href{mailto:weiru.liu@bristol.ac.uk}{weiru.liu@bristol.ac.uk}
}
}

\maketitle

\begin{abstract}
    In this paper, we adopt constrastive explanations within an end-user application for temporal planning of smart homes.
    In this application, users have requirements on the execution of appliance tasks, pay for energy according to dynamic energy tariffs, have access to high-capacity battery storage, and are able to sell energy to the grid.
    The concurrent scheduling of devices makes this a multi-effector planning problem, while the dynamic tariffs yield costs that are non-stationary (alternatively, costs that are stationary but depend on exogenous events).
    These characteristics are such that the planning problems are generally not supported by existing PDDL-based planners, so we instead design a custom domain-dependent planner that scales to reasonable appliance numbers and time horizons.
    We conduct a controlled user study with 128 participants using an online crowd-sourcing platform based on two user stories.
    Our results indicate that
    users provided with contrastive questions and explanations have higher levels of satisfaction, tend to gain improved understanding, and rate the helpfulness more favourably with the recommended AI schedule compared to those without access to these features.
\end{abstract}


\section{Introduction}
In a major survey for the explainable AI (XAI) community on relevant literature from social science, Miller~\cite{Miller:AIJ:2019} advocated an important theory on human explanations: that they are answers to contrastive why-questions of the form \emph{why P rather than Q?} called contrastive explanations~\cite{Lipton:RIPS:1990}.
Although recent interest in XAI has largely focused on machine learning, there has also been parallel interest in XAI for AI planning, referred to as explainable AI planning (XAIP)~\cite{Chakraborti:IJCAI:2020}.
Krarup et al.~\cite{Krarup:JAIR:2021} recently introduced a definition of constrastive explanations for XAIP based on planning model restrictions.
According to this definition, a contrastive question about plan $\pi$ for planning problem $\Pi$ forms a set of constraints on $\Pi$ yielding a restricted hypothetical planning problem $\Pi'$.
A contrastive explanation then is an answer to this question in the form of a comparison between $\pi$ and some hypothetical plan $\pi'$ for $\Pi'$.
For example, a contrastive question ``why is action $A$ used in state $S$, rather than action $B$?'' might have a contrastive explanation ``because using $A$ in $S$ permits the best plan ($\pi$) with lower cost than a plan ($\pi'$) using~$B$''.
A series of such questions and explanations forms an interactive dialogue between user and XAI where the general objective is to help the user better understand the original plan~$\pi$ and/or planning problem $\Pi$.

There is growing evidence that interactiveness is key to the success of XAI tools.
For example, Rong et al.~\cite{Rong:PAMI:2024} conducted a meta-analysis of results from recent XAI user studies and concluded that interactiveness in particular has a positive effect on user trust, satisfaction, and comprehension.
Indeed the work by Krarup et al.\ is closely related to an established concept in AI planning that relies heavily on interactiveness, known as planning as an iterative process~\cite{Smith:AAAI:2012}.
In this setting, AI planning serves as a decision-support tool for requirements elicitation; it assumes that requirements are not fully specified but through an iterative process of AI planning, plan inspection, and planning model revision, humans can refine the model so as to fully specify their requirements.
Although the objective of constrastive explanations is not necessarily to revise a model, it nonetheless remains a valid use case, e.g.\ given a contrastive explanation based on $\Pi$ and $\Pi'$, a user may decide that $\Pi'$ better reflects their requirements and thus choose to adopt it as their new planning model (becoming the subject of future contrastive questions).
As such, contrastive explanations may be complementary to planning as an iterative process.

In this paper we study contrastive explanations in the context of an end-user application for the scheduling of smart homes, called Cuttlefish.
This application assumes that users pay for energy according to dynamic energy tariffs, have access to high-capacity battery storage, and have the ability to sell energy to the grid.
With dynamic tariffs, such as those offered by Octopus Energy in the UK, prices vary according to wholesale prices in timeslots of e.g.\ 30 minutes each.
This allows consumers to reduce energy costs by shifting energy consumption from when prices are high to when prices are low.
Consumers can further reduce energy costs by augmenting their home energy supply with high-capacity battery storage.
This allows them to store low-cost energy for use when grid prices are high and/or to sell surplus energy back to the grid.
This idea forms the basis of the GridFLEX platform by Powervault and has been explored by previous work~\cite{McAreavey:ICAART:2022}.
Beyond this idea, however, Cuttlefish further assumes that users have requirements on home appliance tasks where there is some temporal flexibility in their execution (e.g.\ washer cycles).
The overall objective of Cuttlefish then is to recommend a home schedule for the upcoming week---for the battery and for all appliances---so as to satisfy user requirements while minimising energy costs.
This is treated as a multi-effector temporal planning problem.
The inclusion of contrastive explanations in Cuttlefish is intended to help users understand why particular schedules are being recommended, and potentially to support users in revising their requirements.

Constrastive explanations rely on successive calls to an AI planner; a design referred to by Krarup et~al.~\cite{Krarup:JAIR:2021} as XAIP-as-a-service.
The ability to solve planning problems in reasonable time is thus crucial to achieving an acceptable user experience in Cuttlefish, both to generate the original recommendations and to answer contrastive questions.
An obvious approach is to compile problems into PDDL~\cite{Edelkamp:TR:2004} and then use an off-the-shelf domain-independent planner.
Unfortunately this approach is not viable due to certain characteristics of the Cuttlefish domain, especially non-stationary costs that are a consequence of dynamic tariffs.
These characteristics are similar to certain other real-world planning problems that have been previously studied in the AI planning literature, including unit commitment~\cite{Campion:ICAPS:2013,Piacentini:ICAPS:2016} and electrical balancing~\cite{Piacentini:ICAPS:2013}.
To model these problems, an unofficial variant of PDDL introduced a language feature known as timed initial fluents~\cite{Coles:JAIR:2012}, generalising an earlier language feature from PDDL 2.2 known as timed initial literals~\cite{Edelkamp:TR:2004}.
Both language features are used to model unconditional exogenous events that determine the values of a subset of state variables where timed initial fluents apply to numeric variables and time initial literals to Boolean variables.
With timed initial fluents, non-stationary costs in Cuttlefish can be understood as stationary costs that depend on state variables whose values are determined by exogenous events.
However, support for time initial fluents remains limited among domain-independent planners; to the best of our knowledge, this only applies to \textsc{UPMurphi}~\cite{DellaPenna:ICAPS:2009} and \textsc{Popf}~\cite{Coles:ICAPS:2010} as well as some extensions such as \textsc{DiNo}~\cite{Piotrowski:IJCAI:2016} and \textsc{Popf-Tif}~\cite{Piacentini:AAAI:2015}.
More importantly, while planners may support timed initial fluents, none appear to support costs that depend on the affected state variables.
This implies that Cuttlefish planning problems are not solvable by existing domain-independent planners.

To enable Cuttlefish we develop our own state-space search planner using A$^{*}$ coupled with a domain-dependent heuristic.
Our implementation scales to reasonable time horizons and appliance numbers, equating to a one-week schedule divided into one-hour timeslots with up to four concurrent appliances.
We use a microservice architecture that consists of a frontend web interface, a web API, and a multi-threaded backend service.
The web API accepts planning problems from the frontend, with previously unseen problems added to a processing queue, and returns solutions for problems that have already been processed.
The backend service dequeues problems and solves them, one problem per thread, using the domain-dependent planner with a fixed budget on runtime and visited states.
The frontend allows users to view recommended schedules and to ask contrastive questions where contrastive explanations are displayed as a comparison between the recommended schedule $\pi$ and an alternative schedule $\pi'$.

We conduct a controlled study with 128 participants via Prolific, containing a treatment group (with contrastive explanation) and a control group (without contrastive explanation). 
The study simulate two scenarios with varying requirements, representing different user profiles. 
We evaluate our XAI approach using quantitative analysis of Likert-scale responses and qualitative analysis of free-text feedback from a questionnaire. 
Results indicate that users with access to contrastive questions and explanations demonstrate higher levels of satisfaction with the recommended schedule.
Moreover, the users tend to show improved understanding of the schedule and rate its helpfulness more favourably.

The rest of the paper is organised as follows:
in Section~\ref{sec:methodology} we present our methodology, including the problem formulation and study design;
in Section~\ref{sec:results} we present our results;
and in Section~\ref{sec:conclusion} we reflect on insights and present conclusion.

\section{Methodology}\label{sec:methodology}

\subsection{Preliminaries}
We consider a variation on classical planning where applicable actions, transitions, and costs may vary over time and where plans have a fixed number of actions.
A non-stationary finite-horizon planning problem, or an NF planning problem, is a tuple $\Pi = (S, A, h, E, T, C, G, s_{1})$ where 
$S$ is a set of states, 
$A$ is a finite set of actions,
$h \in \mathbb{N}$ is a horizon, 
$E = \{ E_{t} \mid t = 1, \dots, h \}$ with $E_{t} : S \to 2^{A}$ the applicable action function at timestep $t$,
$T = \{ T_{t} \mid t = 1, \dots, h \}$ with $T_{t} : S \times A \to S$ the transition function at timestep $t$, 
$C = \{ C_{t} \mid t = 1, \dots, h \}$ with $C_{t} : A \to \mathbb{R}$ the cost function at timestep $t$, 
$G \subseteq S$ is a goal, 
and $s_{1} \in S$ is an initial state.
We say that $E$ is stationary if $E_{1} = \dots = E_{h}$, otherwise $E$ is non-stationary.
By abuse of notation, if $E$ is stationary then we treat $E$ as a function rather than a set.
This terminology and notation extends to $T$ and $C$ similarly.
We say that $\Pi$ is stationary if $E$, $T$, and $C$ are stationary, otherwise $\Pi$ is non-stationary.
A sequence of actions $\pi \in A^{h}$ is a plan if for each $t = 1, \dots, h$ with $s_{t+1} = T_{t}(s_{t}, \pi_{t})$ then $\pi_{t} \in A_{t}(s_{t})$ and $s_{h+1} \in G$.
The cost of a plan $\pi$ is $C(\pi) = \sum_{t = 1}^{h} C_{t}(\pi_{t})$.
A plan is optimal if there is no other plan with lower cost.

\subsection{Problem formulation}\label{sec:problem}
We consider a home consisting of a single battery and zero or more appliances where the energy bill is determined by dynamic tariffs for importing and exporting of energy.
When the battery discharges, the energy is consumed by any active appliance tasks while any surplus energy is exported to the grid.
The objective is to find an optimal schedule for the home over a fixed time period that satisfies some user requirements on appliance tasks.
We model this as an NF planning problem.

\begin{definition}\label{def:tariff}
    A dynamic energy tariff is a tuple $(h, P_{I}, P_{E})$:
    $h \in \mathbb{N}$ is the horizon,
    $P_{I} : H \to \mathbb{R}$ is the import tariff,
    $P_{E} : H \to \mathbb{R}$ is the export tariff,
    where $H = \{ 1, \dots, h \}$ is the set of timesteps.
\end{definition}

Definition~\ref{def:tariff} permits negative import and export prices for any timestep $t$ such that $P_{I}(t) < 0$ means the home is paid to import energy from the grid while $P_{E}(t) < 0$ means the home is charged to export energy to the grid.
Such prices are not necessarily permitted by real dynamic tariffs.
The Agile tariff from Octopus Energy satisfies $P_{I}(t) \le 35$p/kWh and $P_{E}(t) \ge 0$p/kWh for each $t$.

\begin{definition}\label{def:battery}
    Let $(h, P_{I}, P_{E})$ be a dynamic energy tariff.
    A battery specification is a tuple $(\omega, \lambda)$:
    $\omega \in \mathbb{N}$ is the capacity in timesteps,
    $\lambda \in \mathbb{R}^{+}$ is the (dis)charge rate per timestep,
    where $\omega \cdot \lambda$ is the energy capacity.
    A battery scheduling model is an NF planning model $\Pi$:
    \begin{itemize}
        \item $S = \{ 0, 1, \dots, \omega \}$,
        $A = \{ -1, 0, 1 \}$ such that for each $s$:
        \begin{LaTeXdescription}
            \item[Discharge] $-1 \in E(s)$ if $s \ne 0$ \quad  $\mathbf{Off}$ $0 \in E(s)$
            \item[Charge] \qquad $1 \in E(s)$ if $s \ne \omega$
        \end{LaTeXdescription}
        \item $T(s, a) = \max\{ 0, \min\{ s + a, \omega \} \}$ for each $s$ and $a$
        \item For each $t$ and $a$:
        \begin{equation}
            C_{t}(a) =
            \begin{cases}
                P_{E}(t) \cdot a \cdot \lambda & \text{if } a = -1 \\
                P_{I}(t) \cdot a \cdot \lambda & \text{otherwise}
            \end{cases}
        \end{equation}
    \end{itemize}
\end{definition}

According to Definition~\ref{def:battery}, the battery serves as a finite-but-replenishable resource allowing homes on dynamic energy tariffs to earn by discharging to the grid.
Due to Definition~\ref{def:tariff}, the costs in Definition~\ref{def:battery} are non-stationary and may be positive or negative.
Note that Definition~\ref{def:battery} is equivalent to the battery scheduling model from~\cite{McAreavey:ICAART:2022} but excluding the predicted home energy consumption parameter, or alternatively assuming that these predictions are always zero.
The state space in Definition~\ref{def:battery} tracks the battery charge level as represented by non-negative integers.
This choice simplifies implementation (e.g.\ by avoiding floating-points) but has no conceptual significance.

\begin{definition}\label{def:appliance}
    Let $(h, P_{I}, P_{E})$ be a dynamic energy tariff.
    An appliance specification is a tuple $(\omega, \lambda, R)$:
    \begin{itemize}
        \item $\omega \in \mathbb{N}$ is the task duration in timesteps
        \item $\lambda \in \mathbb{R}^{+}$ is the consumption rate per timestep
        \item $R \subseteq 2^{H} \times \mathbb{N}$ is a set of primitive requirements
    \end{itemize}
    where $\omega \cdot \lambda$ is the energy consumption per task, $h$ is the horizon.
    An appliance scheduling model is an NF planning model $\Pi$:
    \begin{itemize}
        \item $S = \{ 0, 1, \dots, \omega \} \times \left( \mathbb{N}_{0} \right)^{|R|}$
        \item $A = \{ 0, 1 \}$ such that for each $t$ and $s$:
        \begin{LaTeXdescription}
            \item[Off] $0 \in E_{t}(s)$ if $s_{(1)} \in \{ 0, \omega \}$
            \item[On] $1 \in E_{t}(s)$ if $\omega - s_{(1)} \le h - t$
        \end{LaTeXdescription}
        \item $T_{t}(s, a) = s'$ for each $s$ and $a$ such that:
        \begin{equation}
            s_{(1)}' = \begin{cases}
                s_{(1)} + 1 & \text{if } s_{(1)} \ne \omega \text{ and } a = 1 \\
                a & \text{otherwise}
            \end{cases} \label{eq:appliance-state-active}
        \end{equation}
        and for each $r_{i} \in {P}$:
        \begin{equation}
            s_{(2, i)}' = \begin{cases}
                s_{(2, i)} + 1 & \text{if } s_{(1)} = \omega \text{ and } H' \subseteq r_{(i, 1)} \\
                s_{(2, i)} & \text{otherwise}
            \end{cases} \label{eq:appliance-state-completed}
        \end{equation}
        where $H' = \{ t - \omega, \dots, t \}$
        \item For each $t$ and $a$:
        $C_{t}(a) = P_{I}(t) \cdot a \cdot \lambda$
        \item $G = \{ s \in S \mid \forall r_{i} \in R, s_{(2, i)} \ge r_{(i, 2)}  \}$
        \item $s_{1} = (0, \dots, 0)$
    \end{itemize}
\end{definition}

According to Definition~\ref{def:appliance}, appliance tasks are fixed-length durative actions (i.e.\ sequences of \emph{on} actions of length $\omega$).
For homes on dynamic energy tariffs with requirements over tasks (e.g.\ on how many to execute and when), tasks can be scheduled to minimise costs while ensuring those requirements are satisfied.
Note that a primitive requirement in Definition~\ref{def:appliance} is a pair $(H', m)$ where $m$ is a minimum number of tasks required to satisfy the requirement and $H'$ is a set of timesteps during which the tasks should be executed.
The state variable $s_{(1)}$ in Definition~\ref{def:appliance} tracks the number of completed steps during an active task, while the state variable $s_{(2, i)}$ for primitive requirement $r_{i}$ tracks the number of completed tasks associated with that requirement.
The applicable actions in Definition~\ref{def:appliance} determine  when a task can be started and stopped; once a task is started it must continue to completion.
The goal in Definition~\ref{def:appliance} requires that all requirements are satisfied, meaning that requirements are treated as hard constraints.

\begin{definition}\label{def:home}
    Let $(h, P_{I}, P_{E})$ be a dynamic energy tariff, $\Pi_{b}$ be a battery scheduling model, and $\Pi_{1}, \dots, \Pi_{n}$ be $n$ appliance scheduling models.
    A home scheduling model is an NF planning model $\Pi$:
    \begin{itemize}
        \item $S = S_{b} \times S_{1} \times \dots \times S_{n}$
        \item $A = A_{b} \times A_{1} \times \dots \times A_{n}$ such that $E_{t}(s) = E_{b, t}(s) \times E_{1, t}(s) \times \dots \times E_{n, t}(s)$ for each $t$ and $s$
        \item $T_{t}(s, a) = s'$ for each $t$, $s$, and $a$ such that:
        \begin{LaTeXdescription}
            \item[Battery] $s'_{b} = T_{b, t}(s_{b}, a_{b})$
            \item[Appliances] $s'_{i} = T_{i, t}(s_{i}, a_{i})$ for each $i$
        \end{LaTeXdescription}
        \item For each $t$ and $a$:
        \begin{align}
            C_{t}(a) & =
            \begin{cases}
                P_{E}(t) \cdot U(a) & \text{if } U(a) < 0 \\
                P_{I}(t) \cdot U(a) & \text{otherwise}
            \end{cases}\label{eq:joint-cost} \\
            U(a) & = a_{b} \cdot \lambda_{b} + \sum_{i = 1}^{n} \left( a_{i} \cdot \lambda_{i} \right) \label{eq:joint-consumption}
        \end{align}
        \item $G = G_{b} \times G_{1} \times \dots \times G_{n}$
        \item $s_{1} = \left( s_{(b, 1)}, s_{(1, 1)}, \dots, s_{(n, 1)} \right)$
    \end{itemize}
\end{definition}

According to Definition~\ref{def:home}, homes on dynamic energy tariffs can reduce energy costs through a combination of battery scheduling as in Definition~\ref{def:battery} and the scheduling of $n$ appliances as in Definition~\ref{def:appliance}.
In this way, the battery and appliances can be scheduled to make best use of the dynamic energy tariff while ensuring requirements over appliance tasks are satisfied.
Note that in Definition~\ref{def:home} the battery and each appliance occupy disjoint dimensions of the state space.
However, Equations~\ref{eq:joint-cost}--\ref{eq:joint-consumption} mean that in general the cost of a (joint) plan is not decomposable.
This makes Definition~\ref{def:home} a multi-effector (or coordinated multi-agent) planning problem.
Definition~\ref{def:home} can be seen as a variation on the battery scheduling model from~\cite{McAreavey:ICAART:2022} where the predicted home energy consumption is replaced with a decision problem about when to consume energy.

\subsection{Contrastive explanations}
The semantics of appliance scheduling models can also be understood by the following formalism.

\begin{definition}\label{def:requirements}
    Let $\Psi = (\omega, \lambda, R)$ be an appliance specification and $A = \{ 0, 1 \}$ be the set of appliance actions.
    Then $\Psi$ is satisfied by a sequence of appliance actions $\pi \in A^{h}$, denoted $\Psi \models \pi$, iff there exists some $X_{1}, \dots, X_{m} \subseteq H$ such that:
    \begin{itemize}
        \item $X_{1} \cap \dots \cap X_{m} = \emptyset$
        \item $\forall X_{i}, \exists t \in H, X_{i} = \{ t, \dots, t + \omega \}$
        \item $\forall X_{i}, \forall t \in X_{i}, \pi_{t} = 1$
        \item $\forall r = (r_{(1)}, r_{(2)}) \in R, $
        $$|\{ i = 1, \dots, m : X_{i} \subseteq r_{(1)} \}| \ge r_{(2)}$$
    \end{itemize}
\end{definition}

\begin{corollary}\label{cor:requirements}
    Let $(\omega, \lambda, R)$ be an appliance specification and $\Pi$ be its appliance scheduling model.
    Then $\pi$ is a plan for $\Pi$ iff $(\omega, \lambda, R) \models \pi$ for each $\pi \in A^{h}$.
\end{corollary}

In Definition~\ref{def:requirements}, each $X_{i}$ represents the set of timesteps of a task.
The definition thus says that a sequence of appliance actions $\pi$ satisfies an appliance requirement $(H', m)$ if $\pi$ has at least $m$ disjoint subsequences of length $\omega$, consisting only of \emph{on} actions, occuring at timesteps from $H'$.
Also, $\pi$ satisfies a set of appliance requirements $R$ if $\pi$ satisfies each requirement from $R$.
Corollary~\ref{cor:requirements} then says that this formalism is equivalent to a plan for the appliance scheduling model from Definition~\ref{def:appliance}.
According to Krarup et al.~\cite{Krarup:JAIR:2021}, a contrastive question about plan $\pi$ for planning problem $\Pi$ forms a set of contraints on $\Pi$ yielding a restricted planning problem $\Pi'$.
Formally, an NP planning problem $\Pi'$ is a restriction to an NP planning problem $\Pi$ if every plan for $\Pi'$ is also a plan for $\Pi$.

\begin{corollary}\label{cor:restriction}
    Let $(\omega, \lambda, R)$ be an appliance specification and $R'$ be a set of primitive requirements.
    Then $(\omega, \lambda, R \cup R') \models \pi$ iff $(\omega, \lambda, R) \models \pi$ and $(\omega, \lambda, R') \models \pi$ for each $\pi \in A^{h}$.
\end{corollary}

Corollary~\ref{cor:restriction} implies that the addition of appliance requirements to an existing appliance scheduling model $\Pi$ yields a new appliance scheduling model $\Pi'$ that is a restriction to $\Pi$.
This is the basis of contrastive explanations in Cuttlefish.

\begin{definition}\label{def:explanation}
    Let $\Psi = (\omega, \lambda, R)$ be an appliance specification and $R'$ be a set of primitive requirements.
    Then $\langle \pi, \pi' \rangle$ is a contrastive explanation iff $(\omega, \lambda, R) \models \pi$ and $(\omega, \lambda, R \cup R') \models \pi'$ where $\langle (\Pi, \pi), R' \rangle$ is the contrastive question.
\end{definition}

Definition~\ref{def:explanation} says that, in Cuttlefish, a contrastive question about appliance schedule $\pi$ for appliance scheduling model $\Pi$ consists of a set of appliance requirements $R'$ that are added to $\Pi$ yielding a restricted model $\Pi'$.
Then a contrastive explanation is a pair $\langle \pi, \pi' \rangle$ where $\pi$ is the original schedule for $\Pi$ and $\pi'$ is an alternative schedule for $\Pi'$.
Typically we assume that $\pi$ and $\pi'$ are optimal.
This definition then extends to $n$ appliances in the obvious way.
Note that we do not permit contrastive questions about the battery directly, since the battery is scheduled with respect to appliance requirements rather than its own requirements.

\subsection{Software Architecture}
The software architecture for Cuttlefish consists of three components outlined below.
Energy price data was taken from the Agile tariff by Octopus Energy for the London region during the period 11--17 November 2019.
This data is based on 30-minutes timeslots, which were mapped to 1-hour timeslots by ignoring the second datapoint each hour, e.g.\ 00:00--00:30 was mapped to 00:00--01:00, while 00:30--01:00 was ignored.
All import prices were non-negative (min: 5.88, max: 35, mean: 11.46, SD: 5.94), as were export prices (min: 3.96, max: 17.68, mean: 6.51, SD: 2.77).
The domain-dependent planner was based on a direct implementation of the model from Definition~\ref{def:home} except that costs were normalised to be non-negative.
The planner uses a custom implementation of A$^{*}$ with a domain-dependent heuristic.
The heuristic was chosen based on empirical observation of performance on the chosen price data, and primarily serves to prune states from which the goal is unreachable based on straightforward domain-dependent properties.
Full details can be found in the source code.
The three components are as follows:
\begin{itemize}[leftmargin=*, noitemsep, topsep=0pt]
    \item \textbf{cuttlefish-frontend} is the user interface, accessed via a desktop web browser.
    It is implemented in JavaScript using React and various libraries including Chart.js.
    \item \textbf{cuttlefish-api} is a web API that integrates with the database and provides all backend functionality required by the user interface.
    It is implemented in Python using the Falcon web framework and other libraries.
    \item \textbf{cuttlefish-backend} is a system service implemented in Rust that monitors the database for new planning problems, attempts to solve those problems using the (integrated) planner, and records results in the database.
    A thread pool is used to process multiple planning problems concurrently (one per thread).
    For each problem, the planner has a fixed budget on runtime and visited states.
\end{itemize}
During the study cuttlefish-frontend and cuttlefish-api were served through an Nginx reverse proxy.
The thread pool was allocated 12 threads, while the planner was allocated a budget of 3 minutes and 8m states per problem.

\subsection{Study Design}
This section outlines the setup of our study. 
Participants were randomly assigned to the control group or the treatment group, denoted as \texttt{CG} and \texttt{TG} respectively; each presented with a different user interface: the \texttt{CG} only reviewed the recommend schedule, while the \texttt{TG} reviewed the recommend schedule, had options to ask questions, and received explanations.

Two characters were created to simulate diverse requirements, named Alice (\texttt{A}) and Bob (\texttt{B}). 
\texttt{A} represents a working professional (working hours 9:00--17:00), while \texttt{B} represents a night-shift worker (23:00-9:00). 
\texttt{B}'s schedule imposes more restrictions on time windows compared to \texttt{A}'s, leading to increased complexity. 
For example, \texttt{A} has 2 days working from home, \texttt{B} has none; \texttt{A} uses 1 washer-dryer cycle per week, \texttt{B} uses 2 with once by Thurs. and once after.
This design serves our objective of testing the potential of the XAI interface in differing requirements. 
The appliance specifications $(\lambda, \omega)$ are chosen as follows: washer $(2, 0.75)$, dryer $(3, 1.50)$, dishwasher $(1, 1.20)$, vehicle $(4, 5.00)$ and battery $(6, 1.00)$.
Participants follow the workflow below:
\begin{itemize}[leftmargin=*, noitemsep]
    \item \textbf{Application description:}
    Participants are provided with domain knowledge including information on dynamic energy tariffs, home battery, and four smart home appliances with images and animations for illustration. 
    A brief paragraph outlines the tasks assigned to the participants.

    \item \textbf{User manual:}
    Participants are presented with a manual on how to use the interface. 
    Participants in \texttt{CG} receive instructions only on reviewing schedules. 
    Participants in \texttt{TG} receive additional information on asking contrastive questions, with examples provided. 
    Based on feedback from a pilot study, a small exercise is incorporated to familiarise users with adding appliance requirements.

    \item \textbf{Step I (\texttt{A}):}
    Participants read \texttt{A}'s \emph{user story}, including the appliances and requirements. 
    They are redirected to the main page where the \emph{schedule chart} of the recommended schedule $\pi$ is shown. 
    In \texttt{CG}, participants can optionally view the price chart and \emph{recommended schedule chart} before proceeding to the next stage.
    Participants in \texttt{TG} can submit questions about the recommended schedule via \emph{question feature}.
    The \emph{contrastive schedule charts} appear after the planning problem is solved. 
    Charts of the recommended schedule and alternative schedule are presented with \emph{explanation feature}. 
    Participants in \texttt{TG} were required to ask at least five different questions before proceeding to the next stage.
    
    \item \textbf{Step II (\texttt{B}):}
    This step follows the same procedure as \textbf{Step I} with \texttt{B}'s \emph{user story} and a different recommended schedule.

    \item \textbf{Questionnaire:}
    The questionnaire begins with three questions about their previous experience with smart home devices, educational background, and AI experience. 
    We use a 7-point Likert scale (1-7) to measure opinions, with higher values indicating stronger agreement. 
    Partially inspired by \cite{Eifler:ICAPS:2022}, common questions across both groups cover perceptions difficulty, satisfaction, and usefulness of schedules for two characters.
    Some questions vary between groups:
    Participants in \texttt{TG} answer two additional questions about the explanations, and are asked about difficulties encountered during the study.
    In \texttt{CG}, the question on difficulties is replaced by a question about the explanations participants would like to have.
    The three key questions are:
    \raisebox{.5pt}{\textcircled{\raisebox{-.9pt} {1}}} In \texttt{A}'s (\texttt{B}'s) case, I find it easy to understand the Cuttlefish AI schedule.
    \raisebox{.5pt}{\textcircled{\raisebox{-.9pt} {2}}} In \texttt{A}'s (\texttt{B}'s) case, how satisfied are you with the Cuttlefish AI schedule?
    \raisebox{.5pt}{\textcircled{\raisebox{-.9pt} {3}}} I think the Cuttlefish AI schedule will be useful for \texttt{A} (\texttt{B}).
\end{itemize}

\begin{figure}[!t]
    \begin{subfigure}[c]{0.49\columnwidth}
    \centering
    \includegraphics[trim={4.5cm 1cm 4.5cm 1cm},clip,height=4.5cm]{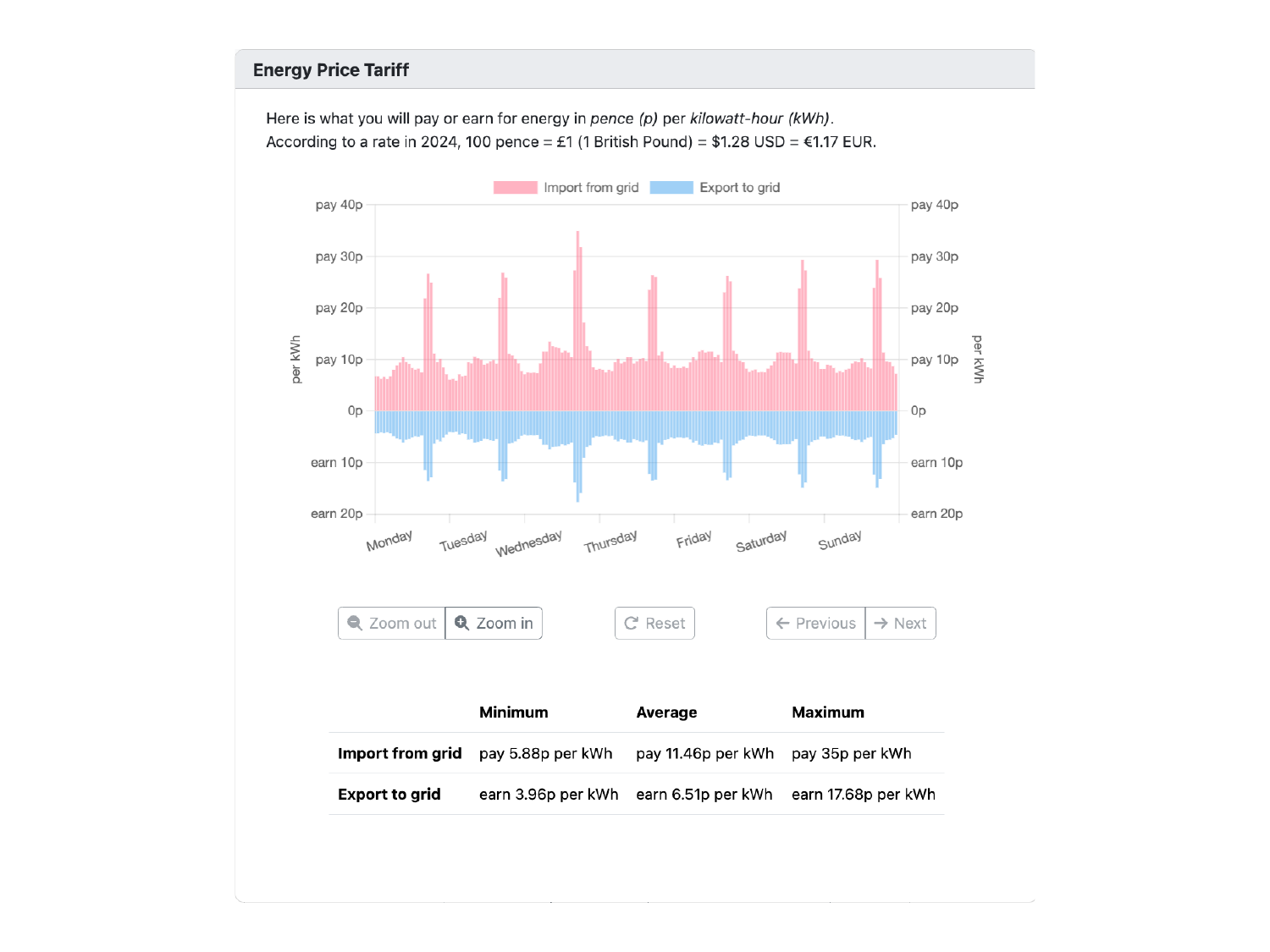}
    \caption{Price chart.}
    \label{fig:interface_screenshot_price}
    \end{subfigure}
    \hfill
    \begin{subfigure}[c]{0.49\columnwidth}
    \centering
    \includegraphics[trim={4cm 0.8cm 4cm 0cm},clip,height=4.5cm]{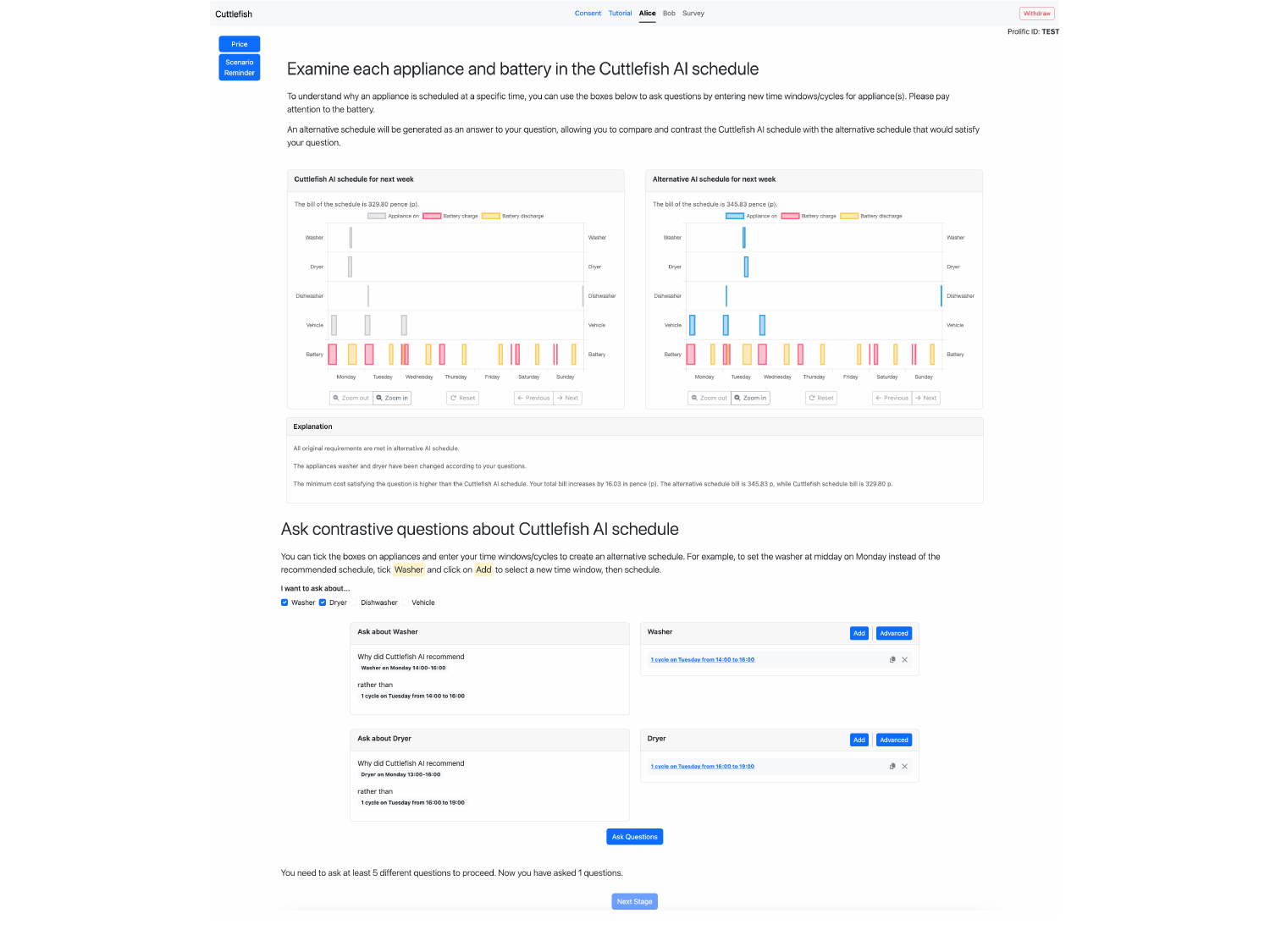}
    \caption{Main page.}
    \label{fig:interface_screenshot_main}
    \end{subfigure}
    \caption{Screenshots of the interface. On the main page, rows are schedule charts, explanations and questions.}
    \label{fig:screenshots}
\end{figure}

\subsection{User Interface (Treatment Group)}
The Cuttlefish web interface consists of the price chart (Figure \ref{fig:interface_screenshot_price}), the schedule charts, and the XAI sections (Figure \ref{fig:interface_screenshot_main}, including question and explanation features).

\textbf{Price chart} displays the dynamic price tariff in one-hour slots over a week. 
The purpose of the price chart is to inform users about price variations, helping them identify preferable time windows for appliance tasks. 
Users can access this chart by selecting a floating button on the left side of the interface.

\textbf{Recommended schedule chart} presents the optimal schedule $\pi$ under the primitive model $\Pi$.
The chart has five rows, representing the tasks of four appliances (washer, dryer, dishwasher, and vehicle) and the battery with different colours. 
At the top of each chart, users can view the corresponding cost of the schedule. 
The recommended schedule chart supports interactive features such as zooming in/out and provide detailed information when user selects a task.
Both \texttt{CG} and \texttt{TG} participants can view the recommended schedule chart.
\textbf{Contrastive schedule charts} appear only after a contrastive question is asked for \texttt{TG}, presenting two schedules for comparison: the optimal schedule $\pi$ under the primitive model $\Pi$ (left) and a schedule $\pi'$ under the contrastive model $\Pi'$ (right). 

\textbf{Question feature}
locates below the schedule charts which allows users to ask contrastive questions about appliances. 
To set up a restricted planning problem $\Pi'$, users can select checkboxes for targeted appliances and input new time windows and cycles. 
The system then generates a new schedule $\pi'$ under these modified constraints and visually displays the updated contrastive schedule chart. 
The accompanying explanation section provides contrastive schedule charts and contrastive (textual) explanations on the cost to users.

\textbf{Explanation feature}
provides the contrastive explanation of two possible situations:
\begin{itemize}[leftmargin=*, noitemsep]
    \item 
        The new problem is solvable. \emph{``The minimum cost satisfying the question is higher than the Cuttlefish AI schedule. Your total bill increases by $C(\pi')-C(\pi)$ in pence (p).''}
    \item 
        The new problem is unsolvable, exceeded time limit, memory or space budget. \emph{``Unsolvable problem / Memory budget exceeded / Space budget exceeded. Please adjust your question and try again.''}
\end{itemize}

\subsection{Recruitment}

Participants were recruited using Prolific.
Base compensation was according to an hourly rate of \pounds 8.
The study was advertised with a suggested completion time, and participants were compensated according to the median time taken of the group.
Participants who exceeded the suggested study time received a \pounds 1 bonus.
All participants were recruited without being informed of their group assignment.
To ensure data quality, several recruitment filters were applied prior to the study using tools by Prolific.
Potential participants were required to be over 18 years old, proficient in English, and have a good record on Prolific.
We set criteria of a minimum 75\% approval rate and completion of 100 studies.
After the study, we filtered out participants who had completed the study exceptionally quickly ($<$ 5 minutes), failed more than two attention checks (each indicates inactivity without any interaction for 5 minutes), or demonstrated a clear lack of effort (incomplete or repetitive answers to most questions).

In total, 65 individuals participated in \texttt{TG} and 69 in \texttt{CG}.
After excluding participants based on our criteria, we retained the data from $N$=63 from \texttt{TG} and $N$=65 from \texttt{CG} participants for analysis. 
The final samples included 59 female and 69 male participants. 
Mean age of all participants was 30.63 years (min: 20, max: 74, SD: 10.19).
Of all participants, 44 (34\%) reported previous experience with smart energy meters or dynamic energy tariffs, while 61 (47\%) reported to have a STEM background.
Collected data included questionnaire, the questions asked in each scenario and participation time.
The median time spent on the study was 55 (resp. 15) minutes for \texttt{TG} (resp. \texttt{CG}). 
This discrepancy was primarily due to the additional computation time in \texttt{TG}, where participants spent time waiting for planner responses when a planning problem had not been previously solved and stored in the database. 
Participants in \texttt{TG} asked an average of 7 questions about \texttt{A} and 6.39 questions about \texttt{B}, slightly above the minimum requirement of 5 questions. 
Each contrastive question required an average computation time of 117.5 seconds for the planner.

We conducted an initial pilot study with 30 participants to optimise the user interface design, and an additional 10 participants to estimate participation time on the final interface. 
Data from the pilot studies was not included in final analysis.

\section{Results}\label{sec:results}

We now evaluate key questions and present statistical analysis regarding the understanding, satisfaction, and usefulness of the recommended schedule for the two characters. 
Then we analyse the free-text responses from the questionnaires.

\begin{figure}[!t]
    \centering
    \includegraphics[width=1\columnwidth]{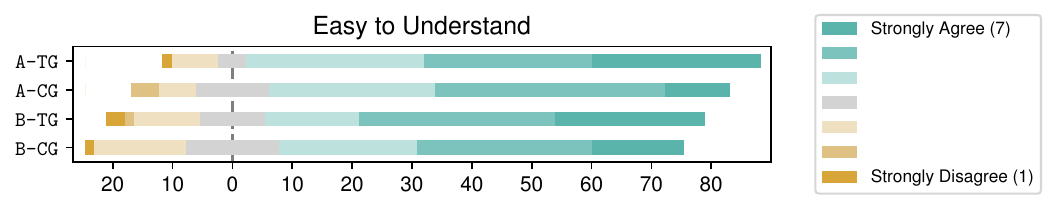}
    \vfill
    \includegraphics[width=1\columnwidth]{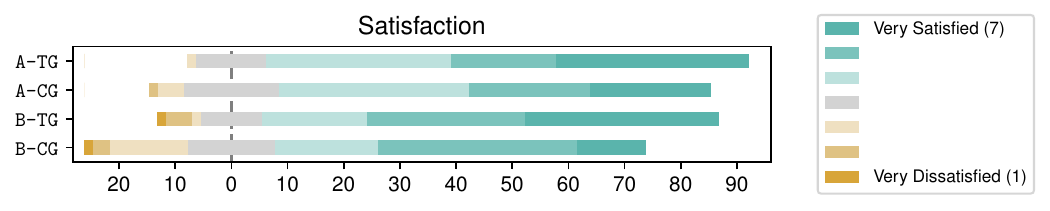}
    \vfill
    \includegraphics[width=1\columnwidth]{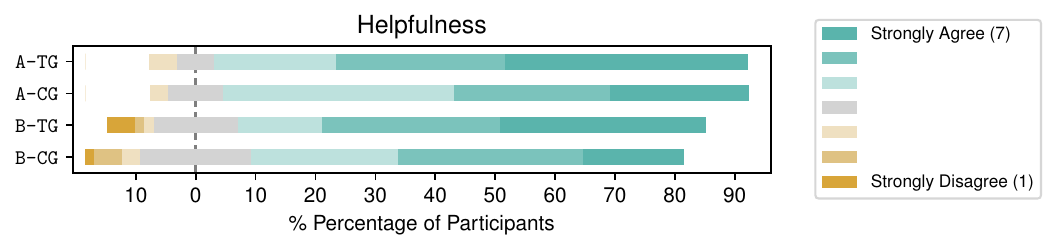}
    \caption{Results of key questions. Abbreviations: Alice \texttt{A}, Bob \texttt{B}, Treatment Group \texttt{TG}, Control Group \texttt{CG}.}
    \label{fig:survey_likert}
\end{figure}

We analyse three key questions answered by both groups in the questionnaire on understanding, satisfaction, and helpfulness to compare the effectiveness of explanations. 
As shown in Figure \ref{fig:survey_likert}, for both characters (\texttt{A} and \texttt{B}), the percentage of participants providing positive feedback (score $>$ 5) is higher in the group with access to questions and explanations (\texttt{TG}) compared to the group without (\texttt{CG}).
This suggests that the contrastive questions and explanations improved user experience with different requirements of our study.
The results indicate that the average scores in \texttt{B} are lower than in \texttt{A}, possibly due to increased complexity.

\subsection {Quantitative analysis on questions}
We conducted Welch's \emph{t}-test on the Likert scale responses to the aforementioned questions. 
We used one-tailed test of the \emph{null} hypotheses $H_0$: the means of \texttt{CG} and \texttt{TG} are equal, and the \emph{alternative} hypotheses $H_1$: the mean of \texttt{CG} is \emph{less} than the mean of \texttt{TG}.
Table \ref{tab:p_values} reports the mean, standard deviation and $p$-value metrics on a Likert scale of the three questions.

The data suggest that participants reported favorable outcomes across all three metrics.
In \texttt{A}'s scenario, there was a significant difference in \emph{satisfaction} and \emph{helpfulness} between the \texttt{TG} and \texttt{CG} ($p$ = \textbf{.033} and \textbf{.029}, respectively.)
We reject the $H_1$ for \emph{understanding} score ($p$ = .055). However, the mean \emph{understanding} score was higher in the \texttt{TG} compared to the \texttt{CG}. 
In \texttt{B}'s scenario, there was a significant difference in \emph{satisfaction} between the two groups ($p$ = \textbf{.01}). 
We obtained $p$ = .169 and $p$ = .076 for \emph{understanding} and \emph{helpfulness} respectively.
In addition to the key questions, we found that participants' views on their understanding of AI's usage in smart homes scored slight higher in the \texttt{TG} (mean: 5.797, SD: 1.416) compared to the \texttt{CG} (mean: 5.569, SD: 1.323) without statistical significance ($p$ = .173).
The \texttt{TG} also demonstrated high confidence in AI through explanations (mean: 5.531, SD: 1.345) and believed in the helpfulness of adopting AI technologies (mean: 6.078, SD: 1.131).

We can conclude that participants in the \texttt{TG} gained significantly more \emph{satisfaction} on the recommended schedule.
The trend suggests that participants in the \texttt{TG} also found the recommended schedules be (i) easier to understand in both scenarios, and (ii) more helpful than those in the \texttt{CG}. 

\subsection{Qualitative Analysis on Texts}
In the free-text question, the participants are asked to 
(a) describe their understanding of the schedule under each character (both \texttt{TG} and \texttt{CG});
(b) describe the most difficult part in understanding the explanations (\texttt{TG} only).
While the free-context questions were diverse in both length and topics, some themes and feedback were captured in the answers.
\begin{table}[t]
    \begin{threeparttable}
    \caption{Statistics of Likert Scales (7) on Key Questions.}
    \label{tab:p_values}
    \begin{tabular}{ccccc}
        \toprule
        \multirow{2}{*}{\textbf{Question}} & \multirow{2}{*}{\textbf{Scenario}} & \multicolumn{2}{c}{\textbf{Mean$\pm$Std}} & \multirow{2}{*}{\textbf{$p$-value}}\\
        \cline{3-4}
        & & \texttt{TG} & \texttt{CG} & \\
        \midrule
        \multirow{2}{*}{\makecell{Easy to \\ Understand}}& \texttt{A} & \textbf{5.578}$\pm$1.307 & 5.215$\pm$1.256 & .055 \\
        & \texttt{B} & \textbf{5.328}$\pm$1.564 & 5.077$\pm$1.395 & .169 \\
        \midrule
        \multirow{2}{*}{Satisfaction} & \texttt{A} & \textbf{5.718}$\pm$1.119 & 5.338$\pm$1.215 & \textbf{.033} \\
        & \texttt{B} & \textbf{5.625}$\pm$1.464 & 5.015$\pm$1.452 & \textbf{.01~} \\
        \midrule
        \multirow{2}{*}{Helpfulness} & \texttt{A} & \textbf{5.937}$\pm$1.139 & 5.569$\pm$1.045 & \textbf{.029} \\
        & \texttt{B} & \textbf{5.578}$\pm$1.572 & 5.200$\pm$1.394 & .076 \\
        \bottomrule
    \end{tabular}
    \begin{tablenotes}
        \tiny
        \item Abbreviations: Alice (\texttt{A}), Bob (\texttt{B}), Treatment Group (\texttt{TG}), Control Group (\texttt{CG}).
        \item Sample size $N$=63 (resp. $N$=65) in \texttt{TG} (resp. \texttt{CG}); significance level $\alpha=.05$. The $t$-statistics for understanding, satisfaction, and helpfulness are -1.607 (-0.963), -1.848 (-2.374), and -1.914 (-1.446) for Alice's (Bob's) scenario.
    \end{tablenotes}
    \end{threeparttable}
\end{table}

\subsubsection{Improvement in comprehension}
In \texttt{CG}, many participants (6 times for \texttt{A} and 13 times for \texttt{B}) questioned the appropriateness of individual task in the recommended schedule. 
Some (3 for \texttt{A} and 2 for \texttt{B}) disagreed with the use of the battery to be optimal, or believed a schedule costs less than the recommend schedule (2 for \texttt{A} and 5 for \texttt{B}). Some participants (5) did not fully understand the recommended schedule.
In contrast, in \texttt{TG}, fewer participants commented on time windows (6 for \texttt{A} and 3 for \texttt{B}). Among these comments, most focused on the dependencies between the washer and dryer rather than the appropriateness of individual tasks.

\subsubsection{Criticisms by treatment group}
The most common criticism of the interface was the runtime (6 for \texttt{A} and 2 for \texttt{B}). 
In some cases, users' questions returned unsolved due to computational budget constraints. 
Some participants (2) commented on the design for asking questions by adding restrictions. 
They intended to ask questions by suggesting time windows of appliances (e.g., ``Why washer runs on some time rather than another time?'') However, such questions usually had to obey the primitive requirements on time windows, which led to unwanted additional new tasks.
This issue could potentially be addressed by improving the forms of questions. 

\subsubsection{Expectation on explanation}
Some participants expressed a desire for more types of contrastive questions and explanations. 
For example, one user asked, ``Why did AI schedule appliances on (any) two consecutive days?''. 
Some participants expected more types of explanations, including interpretations of AI decisions for lay users. 
Two users mentioned wanting to understand ``AI's logic'' or ``reasoning behind AI's choice.'' 
Additionally, users expressed a desire for features to handle usage shifts on unexpected (e.g., power outage) or temporary requirement changes, cost/energy breakdown or the expectation of multiple backup options.

\section{Discussion \& Conclusion}\label{sec:conclusion}
\subsection{Task dependencies}
While most participants in the treatment group acknowledged that the schedule achieved the most economical price according to the requirements, or were convinced after asking questions, concerns remained regarding the dependencies between appliances (e.g., the dryer should operate some time after the washer finishes). This dependency feature was considered during the development but not presented to users, both as a simplification and due to practical considerations of computation time. However, this could confuse users and raise concerns about real-world applications.

\subsection{Conclusion}
In this paper, we have adopted contrastive explanations within an end-user application for temporal planning of smart homes. 
We developed a domain-dependent planner for multi-effector planning problems, along with a web-based interface that generates explanations answering contrastive questions about appliances by allowing users to add additional requirements.
We conducted a user study on our implementation, which demonstrated that our approach can improve users' ratings on understanding, satisfaction, and helpfulness of the schedule. 
Our findings allow us to conclude that our XAI feature can enhance users' satisfaction with and acceptance of the recommended schedule. Future work will include further optimisation of efficiency of the planner and support for additional forms of ``why'' questions.

\bibliographystyle{IEEEtran}
\bibliography{IEEEabrv,ictai2024}

\appendix

\begin{table}[h]
    \begin{threeparttable}
    \caption{Comparison of Alice and Bob's Appliance Requirements}
    \label{tab:design_of_requirements}
    \centering
    \scriptsize
    \begin{tabular}{p{0.18\columnwidth}p{0.34\columnwidth}p{0.34\columnwidth}}
    \toprule
    \textbf{Aspect} & \textbf{Alice} & \textbf{Bob} \\
    \midrule
    Work Hours & 9:00--17:00 & 23:00--9:00 \\
    Commute Time & 2 hours & 1 hour \\
    WFH Days & Wedn. and Fri. & None \\
    \midrule
    \texttt{W} \& \texttt{D} Cycles & 1 & 2 \\
    \texttt{W} \& \texttt{D} Time & Not between 23:00--7:00 & Only between 7:00--15:00 \\
    \texttt{DW} Cycles & 2  & 2 \\
    \texttt{EV} Cycles  & 3  & 2 \\
    \midrule
    \multirow{1}{*}{\makecell{Additional\\requirements}} & \texttt{DW} once by Thurs., once after; \texttt{EV} cycles before office days. & Run appliances once on weekdays, once on weekends. \\
    \bottomrule
    \end{tabular}
    \begin{tablenotes}
        \scriptsize
        \item Abbreviations: Work From Home (WFH), Washer (\texttt{W}), Dryer (\texttt{D}), Dishwasher (\texttt{DW}), Electric Vehicle (\texttt{EV}).
    \end{tablenotes}
    \end{threeparttable}
\end{table}


\begin{table}[h]
\caption{Survey questions for Treatment and Control Group}
\label{tab:survey_questions}
\centering
\scriptsize
\setlength{\tabcolsep}{4pt}
\begin{tabular}{|c|p{0.78\columnwidth}|c|}
    \hline
     & \textbf{Question} & \textbf{Type} \\
    \hline
    \multirow{10}{*}{\makecell{\texttt{TG}\\ \& \\ \texttt{CG}}} 
    & Have you used smart energy meters or dynamic energy tariffs? & \multirow{3}{*}{Yes/No}  \\
    & Do you have a STEM background? & \\
    & Do you have any prior experience in Artificial Intelligence (AI)? &  \\
    \cline{2-3}
    & $^\star$In Alice's (Bob's) case, I find it easy to understand the Cuttlefish AI schedule. & \multirow{4}{*}{\makecell{Likert \\ scale \\ (1)-(7)}} \\
    & $^\star$In Alice's (Bob's) case, how satisfied are you with the Cuttlefish AI schedule? & \\
    & $^\star$I think the Cuttlefish AI schedule will be useful for Alice (Bob). & \\
    & This study had helped me understand how AI works on scheduling in smart homes. & \\
    \cline{2-3}
    & Can you briefly describe your views on the Cuttlefish AI schedule? & \multirow{2}{*}{Free-text} \\
    & Do you have any additional comments or feedback on the interface or the tutorial? & \\
    \hline
    \multirow{3}{*}{\texttt{TG}} 
    & The Cuttlefish AI has provided functions where you can ask the AI about schedules for home appliances and modify them. Do you think this would be helpful in real-world applications? & \multirow{3}{*}{\makecell{Likert \\ scale \\ (1)-(7)}} \\
    & I feel the questions and explanations have enhanced my confidence in AI systems in general. & \\
    & I believe questions and explanations will help the adoption of AI technologies. & \\
    \cline{2-3}
    & What is the most difficult part of asking questions or comprehending the explanations? & Free-text \\
    \hline
    \multirow{2}{*}{\texttt{CG}} 
    & Imagine Cuttlefish AI offered a feature allowing users to ask questions and modify schedules for home appliances using AI. Do you think this would be helpful in real-world applications? & \multirow{1}{*}{\makecell{Likert\\ scale \\ (1)-(7)}} \\
    \cline{2-3}
    & Is there any explanation you would like to have about the AI schedule? & Free-text \\
    \hline
\end{tabular}
\begin{tablenotes}
    \scriptsize
    \item $^\star$Key questions.
\end{tablenotes}
\end{table}

\end{document}